%% file: main_arxiv.tex
\documentclass[11pt,twoside]{article} 

\usepackage{eqnarray,amsmath, bm}
\usepackage[utf8]{inputenc} 
\usepackage[T1]{fontenc}    
\usepackage{booktabs}       
\usepackage{amsfonts}       
\usepackage{nicefrac}       
\usepackage{microtype}      

\usepackage{epsf}
\usepackage{epsfig}
\usepackage{fancyhdr}
\usepackage{graphics}
\usepackage{graphicx}
\usepackage{psfrag}
\usepackage{fullpage}
\usepackage{pdfpages}
\usepackage{colortbl}
\usepackage{tcolorbox}
\usepackage[table]{xcolor}
\usepackage{tabularx}
\usepackage{enumitem}
\usepackage{natbib}

\usepackage{tikz}
\definecolor{cyan}{cmyk}{.3,0,0,0}
\definecolor{LightCyan}{rgb}{0.88,1,1}
\definecolor{Gray}{gray}{0.95}

\usepackage{url}
\usepackage[colorlinks,linkcolor=magenta,citecolor=blue, pagebackref=true]{hyperref}
\renewcommand*{\backrefalt}[4]{%
    \ifcase #1 \footnotesize{(Not cited.)}%
    \or        \footnotesize{(Cited on page~#2.)}%
    \else      \footnotesize{(Cited on pages~#2.)}%
    \fi}

\usepackage{amsthm}
\usepackage{amsmath}
\usepackage{amssymb,bbm}
\usepackage{amsmath,amssymb,amsfonts,bm}
\usepackage{subcaption}
\usepackage{caption}

\usepackage{textcomp}
\usepackage{siunitx}
\usepackage{wrapfig}
\usepackage{algorithmic}
\usepackage{algorithm}
\usepackage{tcolorbox}
\usepackage{graphicx}
\usepackage{multirow}
\usepackage{diagbox}

\usepackage{amsmath,amssymb,amsfonts,bm}
\usepackage{amsmath} 
\usepackage{booktabs}
\usepackage{amssymb}
\usepackage{multirow}

\newcommand{\xbm}{{\bm x}}

\newcommand{\Kbm}{{\bm K}}
\newcommand{\zbm}{{\bm z}}

\def\RR{\mathbb{R}}

\begin{document}

\begin{center}

{\bf{\LARGE{LEAF: A Robust Expert-Based Framework for Few-Shot \\ Continual Event Detection}}}
  
\vspace*{.2in}
{\large{
\begin{tabular}{ccccc}
Bao-Ngoc Dao$^{\star1}$ & Quang Nguyen$^{\star2}$ & Luyen Ngo Dinh$^{\star3}$  & Minh Le$^{\star1}$ & Linh Ngo Van$^{1}$
\end{tabular}
}}

\vspace*{.2in}

\begin{tabular}{ccc}
Hanoi University of Science and Technology$^{1}$ \\
FPT Software AI Center$^{2}$ \\
Moreh Inc.$^{3}$
\end{tabular}

\vspace*{.2in}
\today

\vspace*{.2in}

\newcommand\blfootnote[1]{%
  \begingroup
  \renewcommand\thefootnote{}\footnote{#1}%
  \addtocounter{footnote}{-1}%
  \endgroup
}

\blfootnote{$^\star$ Equal contribution}

\input{content/abstract}

\end{center}

\input{content/intro}
\input{content/background}

\input{content/method}
\input{content/experiment}

\input{content/conclusion}

\newpage

\bibliography{aaai2026}
\bibliographystyle{abbrv}

\end{document}

%% file: content/abstract.tex
\begin{abstract}

Few-shot Continual Event Detection (FCED) poses the dual challenges of learning from limited data and mitigating catastrophic forgetting across sequential tasks. Existing approaches often suffer from severe forgetting due to the full fine-tuning of a shared base model, which leads to knowledge interference between tasks. Moreover, they frequently rely on data augmentation strategies that can introduce unnatural or semantically distorted inputs. To address these limitations, we propose \textbf{LEAF}, a novel and robust expert-based framework for FCED. LEAF integrates a specialized mixture of experts architecture into the base model, where each expert is parameterized with low-rank adaptation (LoRA) matrices. A semantic-aware expert selection mechanism dynamically routes instances to the most relevant experts, enabling expert specialization and reducing knowledge interference. To improve generalization in limited-data settings, LEAF incorporates a contrastive learning objective guided by label descriptions, which capture high-level semantic information about event types. Furthermore, to prevent overfitting on the memory buffer, our framework employs a knowledge distillation strategy that transfers knowledge from previous models to the current one. Extensive experiments on multiple FCED benchmarks demonstrate that LEAF consistently achieves state-of-the-art performance.

\end{abstract}

%% file: content/intro.tex
\section{Introduction}

\emph{Continual Event Detection (CED)} integrates the objectives of \emph{Event Detection} and \emph{Continual Learning}. The goal of event detection is to identify and classify events mentioned in text by recognizing and interpreting lexical triggers within their context \citep{doddington2004automatic}. As a foundational task in natural language processing, event detection supports a broad range of downstream applications, including information extraction and knowledge graph construction \citep{nguyen2023retrieving}. Its importance has led to significant research attention and the development of various modeling strategies in recent years~\citep{chen2015event, sha2018jointly, man2022selecting}. Unlike the traditional static learning paradigm, CED operates within a continual learning framework where data arrives incrementally over time. A central challenge in this setting is \emph{catastrophic forgetting} \citep{mccloskey1989catastrophic, ratcliff1990connectionist}, where newly learned information overwrites previously acquired knowledge, leading to a decline in performance on earlier tasks.

Despite recent advances in CED, its performance under data-scarce conditions remains underexplored. In real-world applications, new event types often emerge continuously, making it \emph{impractical to acquire large amounts of labeled data due to high annotation costs}. As a result, models are frequently required to learn from only a few labeled examples for each new event type. To formalize this setting, \cite{zhang2024continual} introduced the task of \emph{Few-shot Continual Event Detection (FCED)}. The problem is particularly relevant in low-resource languages and specialized domains where dataset collection is a significant challenge~\citep{agrawal2025adaptive}. FCED presents two major challenges: \emph{mitigating catastrophic forgetting} and \emph{achieving effective generalization to new event types from limited supervision}.

Recent work such as HANet~\citep{zhang2024continual} addresses FCED by combining few-shot learning techniques with continual learning, employing full model fine-tuning, a memory buffer, and data augmentation. Despite empirical gains, this approach suffers from several limitations. First, fully fine-tuning all parameters for each new task makes the model susceptible to catastrophic forgetting, as knowledge from previous tasks is overwritten. Although the memory buffer provides partial mitigation, its limited capacity in the few-shot setting hinders its ability to preserve and generalize knowledge from past tasks. Specifically, with only a few stored samples per class, the buffer cannot effectively represent the underlying data distribution of earlier tasks. Second, while augmentation strategies such as token shuffling and dropout are used to enrich the input space and reduce overfitting, these techniques may produce unnatural or semantically altered instances. For example, dropout may remove tokens essential for contextual understanding, and token shuffling can disrupt grammatical structure, undermining the semantic cues necessary for accurate event classification. These challenges highlight the need for more robust FCED methods that can mitigate catastrophic forgetting while enabling effective generalization from limited data.

To address these challenges, we propose \textbf{LEAF} (\underline{L}ow-rank \underline{E}xperts for \underline{A}daptive \underline{F}CED), a novel framework incorporating three key innovations. First, inspired by advances in isolated-parameter continual learning~\citep{wang2022learning, wang2023hierarchical, liang2024inflora, le2025adaptive}, LEAF employs an adaptive mixture of experts architecture built upon LoRA modules. Unlike prior work such as Mixture of LoRA Experts (MoLE)~\citep{wu2024mixture}, LEAF introduces a semantic-aware routing mechanism that dynamically selects experts based on input representations from the base model. This instance-level expert selection not only enables modular memory retention but also reduces knowledge interference, thereby mitigating catastrophic forgetting. Second, to enhance learning in limited-data settings and reduce overfitting, LEAF incorporates contrastive learning with label descriptions. These descriptions provide high-level semantic information about event types, helping the model distinguish between closely related classes even when training examples are scarce. Finally, to further reduce overfitting to the memory buffer, we adopt a two-level knowledge distillation strategy, transferring both feature and prediction knowledge from the previous model to the current one. Extensive experiments demonstrate that LEAF consistently achieves state-of-the-art performance on various FCED benchmarks.
Our contributions can be summarized as follows:
\begin{itemize}
    \item We propose \textbf{LEAF}, a novel expert-based framework for FCED that effectively mitigates catastrophic forgetting and enhances generalization in few-shot setting.
    \item We introduce a semantic-aware routing mechanism over LoRA-based experts, enabling dynamic, instance-specific expert selection. This modular design allows for efficient knowledge retention and reduces interference across tasks.
    \item We incorporate contrastive learning with event label descriptions to enhance model discrimination, especially in low-resource scenarios where closely related event types must be distinguished with minimal labeled data.
    \item We adopt a two-level knowledge distillation strategy that transfers knowledge from the previous model, reducing overfitting to the constrained memory buffer.
    \item Extensive experiments on benchmark datasets show that LEAF consistently outperforms existing baselines and achieves state-of-the-art results in the FCED setting.
\end{itemize}



%% file: content/background.tex
\section{Related Work} \label{sec: background}
\subsection{Few-shot Continual Event Detection}

\emph{Continual Event Detection (CED)} is a subfield of continual learning and event detection that focuses on identifying emerging events from sentences within a continuously evolving data stream~\citep{cao2020incremental, yu2021lifelong, le2024sharpseq}. A central challenge in CED is enabling models to learn new tasks effectively while preserving performance on previously encountered ones. To address this, most existing methods adopt rehearsal-based strategies, which store representative samples from earlier tasks for future retraining replay~\citep{cao2020incremental, liu2022incremental, zhang2024continual}. 

Despite recent progress, real-world CED scenarios often involve the continuous emergence of novel event types, where \emph{obtaining a sufficient number of labeled samples for each new event is impractical}. In such cases, models must operate in incremental settings with only a few annotated examples per new event type (e.g., 10, 5, or even 1). To tackle this, \cite{zhang2024continual} introduced \emph{Few-shot Continual Event Detection (FCED)}, a new task that aims to learn event detection incrementally with limited annotated data. The core challenge in FCED lies in adapting to each new task with scarce supervision while retaining knowledge of previously learned tasks. However, standard CED methods often perform suboptimally in this setting, as they primarily focus on mitigating forgetting and tend to overlook the difficulty of learning a new task from limited samples.

\subsection{Mixture of LoRA Experts}

\begin{figure}
    \centering
    \includegraphics[width=0.5\linewidth]{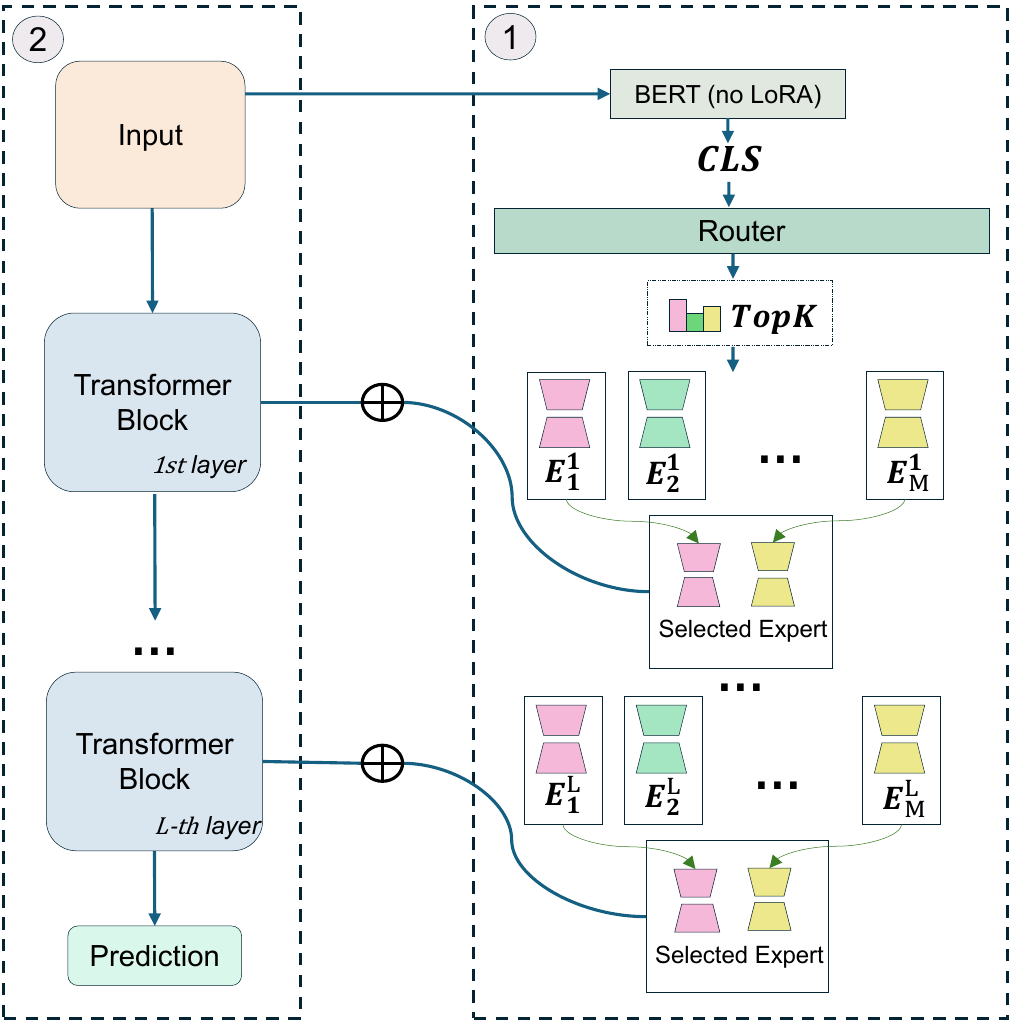}
    \caption{\textbf{LEAF Overall Architecture.} Our framework operates in two stages. In the first stage, the input sentence is passed through the frozen base BERT model to obtain the $[\text{CLS}]$ representation, which is then guides expert selection. In the second stage, the selected LoRA experts are integrated into the base model to compute the final output.}
    \label{fig:model}
\end{figure}

\emph{Mixture of Experts (MoE)} is a class of statistical machine learning frameworks that combine multiple specialized models, or \emph{experts}, to improve model expressiveness and predictive performance~\citep{jacobs1991adaptive}. To enhance computational efficiency, \cite{shazeer2017outrageously} proposed Sparse MoE, which activates only a small subset of experts per input, thereby reducing computational cost. More recently, MoE architectures have been successfully adapted to continual learning settings, demonstrating strong performance in mitigating forgetting and facilitating knowledge transfer~\citep{ yu2021lifelong, chen2023lifelong, le2024mixture}. In parallel, \emph{Low-Rank Adaptation (LoRA)}~\citep{hu2022lora} has emerged as a parameter-efficient technique for fine-tuning pre-trained models by representing weight updates as the product of low-rank matrices, thereby significantly reducing the number of trainable parameters. Building on these advancements, \cite{wu2024mixture} proposed the \emph{Mixture of LoRA Experts (MoLE)}, which integrates LoRA into the MoE framework. In MoLE, each expert is implemented as a distinct LoRA adapter, and a trainable gating network dynamically routes each input token to a selected subset of LoRA experts.

%

%

%% file: content/method.tex
\section{Proposed Method} \label{sec: method}


\subsection{Problem Formulation} \label{sec:problem}

In FCED, a model learns from a sequence of tasks $\{ \mathcal{T}_1, \mathcal{T}_2, \dots, \mathcal{T}_T \}$. Each task $\mathcal{T}_t$ introduces a disjoint set of event labels $\mathcal{R}_t$, such that $\mathcal{R}_i \cap \mathcal{R}_j = \emptyset$ for all $i \ne j$. The training dataset for task $\mathcal{T}_t$ is denoted by $\mathcal{D}_t = \{(\xbm^t_i, y^t_i) \mid y^t_i \in \mathcal{R}_t\}_{i=1}^{ | \mathcal{R}_t | \times K}$, where $K$ is the number of training examples per label, following the \emph{N-way K-shot} learning paradigm~\citep{chen2023consistent}. 

The goal is to enable the model to \emph{rapidly adapt to novel tasks} while \emph{maintaining performance on previously learned ones}. At each task $t$, the model is trained to classify inputs into the event types in $\mathcal{R}_t$, learning new event types without access to data from prior tasks. After completing task $\mathcal{T}_t$, the model is expected to generalize across all previously encountered tasks, accurately identifying event types from the cumulative label set $\widehat{\mathcal{R}}_t = \bigcup_{i=1}^t \mathcal{R}_i$.

Following prior work~\citep{zhang2024continual}, we adopt a memory buffer to mitigate catastrophic forgetting. Given the data scarcity inherent in few-shot learning, we store only a single exemplar per class. Let $\mathcal{M}_i$ denote the set of exemplars stored for the classes introduced in task $\mathcal{T}_i$. At the beginning of task $t$, the accumulated memory from all previous tasks is $\widehat{\mathcal{M}}_t = \bigcup_{i=1}^{t-1} \mathcal{M}_i$. To enhance the utility of these stored examples, we employ the prototype-based data augmentation strategy proposed by~\cite{zhang2024continual}. The final training set for task $t$, denoted by $\widehat{\mathcal{D}}_t$, is constructed by combining the current task data $\mathcal{D}_t$, the accumulated memory $\widehat{\mathcal{M}}_t$, and the corresponding augmented samples.

\subsection{Adaptive LoRA Experts} \label{sec:lora_expert}

In our approach to FCED, we adopt a pre-trained BERT model~\citep{devlin2019bert} as the base model, maintaining consistency with prior work~\citep{zhang2024continual}. To adapt the model to continual learning tasks while preserving efficiency, we integrate LoRA into a modular expert framework.
Previous approaches, such as HANet~\citep{zhang2024continual} typically rely on fine-tuning a single base model across all tasks, which often leads to catastrophic forgetting due to knowledge interference. To mitigate this issue, we propose a modular architecture composed of multiple specialized experts, each parameterized using LoRA.

\input{content/algorithm}

Specifically, for the attention layer in the $l$-th transformer block, we instantiate \emph{a pool of $M$ LoRA experts}, denoted by $\boldsymbol{E}^l = \{E^l_1, \dots, E^l_M\}$. Each expert $E^l_k: \RR^d \rightarrow \RR^{d_{out}}$ is defined by a low-rank parameterization:
\begin{align}
E^l_k(\xbm) = A^l_k B^l_k \xbm, \ A^l_k \in \mathbb{R}^{d_{out} \times r}, \ B^l_k \in \mathbb{R}^{r \times d}, 
\end{align}
where $r \ll \min(d_{\text{out}}, d)$ is LoRA rank, and $\xbm \in \RR^d$ is the input embedding. Each expert $E^l_k$ is associated with a learnable routing vector $W_k^l \in \mathbb{R}^d$, which determines its relevance to a given input.

To select the most appropriate experts for an input sentence $\xbm_i \in \widehat{\mathcal{D}}_t$, we first pass it through the frozen base BERT model to obtain the $[\text{CLS}]$ representation, denoted by $h_{\xbm_i}^{\text{CLS}} \in \mathbb{R}^d$. We then identify the top-$K$ relevant experts by selecting the set of indices $\boldsymbol{K}_{\xbm_i}^l$ that maximizes the sum of their score functions:
\begin{align} 
\boldsymbol{K}_{\xbm_i}^l
= \operatornamewithlimits{argmax}_{S \subset \{1,\dots,M\},\ |S|=K} \sum_{k \in S}  s_k^l(\xbm_i) 
= \operatornamewithlimits{argmax}_{S \subset \{1,\dots,M\},\ |S|=K} \sum_{k \in S}  {W^l_k}^\top  h_{\xbm_i}^{\text{CLS}},
\label{eq:index_selection}
\end{align}
where $s_k^l(\xbm) = {W^l_k}^\top  h_{\xbm}^{\text{CLS}}$ denotes the  score function between expert $E_k^l$ and the input $\xbm$. 

The final output is computed by combining the base model with the selected LoRA experts as follows:
\begin{align}
    h(\xbm) = 
    W_{\text{base}} \xbm
    + \sum_{k \in \Kbm_{\xbm_i}^l}
    \frac{s^l_k(\xbm_i)}{\sum_{k' \in \Kbm_{\xbm_i}^l} s^l_{k'}(\xbm_i)} E^l_k(\xbm)
    ,
\label{eq:update}
\end{align}
where $W_{\text{base}} \in \RR^{d_{out} \times d}$ is the weight from the pre-trained base model. For clarity, $\xbm_i$ refers to the input sentence, while $\xbm$ denotes the input to the $l$-th transformer block during the forward pass of $\xbm_i$.

\input{tables/prompt}

Since the feature vector $h_{\xbm_i}^{\text{CLS}}$ used for expert selection encodes semantic and contextual information, our architecture can dynamically assign specialized experts to distinct regions of the input space. This design allows each expert to focus selectively on the information and patterns required for optimal performance in its specific domain. Functioning as modular memory units, \emph{these experts capture different facets of the input distribution, enabling the model to adaptively activate only a relevant subset of experts for each instance}. This selective routing not only promotes knowledge isolation but also substantially mitigates catastrophic forgetting by minimizing knowledge interference. Unlike prior methods that rely on a single base model to retain all task knowledge~\citep{zhang2024continual}, our approach distributes learning across modular experts, preserving prior knowledge while facilitating the efficient integration of new information. The overall architecture is depicted in Figure~\ref{fig:model}.

Our architecture differs from conventional MoLE frameworks~\citep{wu2024mixture} in that expert selection is performed at the instance level, based on the input sentence, rather than at the token level within each transformer block. By leveraging the $[\text{CLS}]$ representation for expert selection, our model can route inputs based on high-level semantic features, allowing different experts to specialize in distinct semantic domains.

To further enhance expert selection during both training and inference, we introduce a router loss defined as:
\begin{align}
    \mathcal{L}_{\text{router}} = -
    \sum_{\xbm_i \in \widehat{\mathcal{D}}_t}
    \sum_{l}
    \sum_{k \in \Kbm_{\xbm_i}^l}
    s_k^l(\xbm_i),
\end{align}
which encourages the model to maximize the score function of the selected experts.

\subsection{Contrastive Learning with Label Descriptions} \label{sec:label_description}

We define $f(\xbm)$ and $f_{\theta_t}(\xbm)$ as the representations of an input sentence $\xbm$ obtained the base BERT model and the model augmented with LoRA experts, respectively. The parameters of the LoRA experts in the current task $t$, integrated across all transformer blocks, are denoted by $\theta_t$. The final event classification is performed by an MLP event detector, $g_\phi$.

A key challenge in FCED is mitigating the overfitting and learning instability that arise when adapting to new event types with only a limited number of training examples. To mitigate this, we propose incorporating \emph{label descriptions} as auxiliary semantic supervision to enhance generalization and improve robustness.

Specifically, we employ a large language model (LLM), Gemini 2.0~\citep{team2023gemini}, to generate concise, natural-language descriptions for each event label. Examples of these generated descriptions are shown in Table~\ref{tab: instruct}. The resulting descriptions are then encoded using a pre-trained BERT model to obtain their semantic representations. 

As \emph{label descriptions provide high-level semantic information about event types}~\citep{zhang2022efficient}, we aim to align the input representations with the corresponding label semantics. Let $\mathcal{Z}_y$ be the set of representations for the label descriptions associated with event type $y$. We introduce the following contrastive loss to encourage semantic alignment between the input representations and their corresponding label descriptions:
\begin{align}
\mathcal{L}_{\text{label}} &= 
- \sum_{\xbm_i \in \widehat{\mathcal{D}}_t} 
\log 
\frac{\displaystyle
\sum_{\zbm \in \mathcal{Z}_{y_i}} \exp (f_{\theta_t}(\xbm_i) \cdot \zbm)}
{\displaystyle
\sum_{y \in \widehat{\mathcal{R}}_t \setminus {y_i}} \sum_{\zbm' \in \mathcal{Z}_y} \exp(f_{\theta_t}(\xbm_i) \cdot \zbm')}.
\end{align}
This objective encourages the input representation $f_{\theta_t}(\xbm_i)$ to have a higher similarity with the semantic descriptions of its correct event type while minimizing its similarity to descriptions of unrelated types. By aligning inputs with these high-level semantic descriptions, our method provides global information that helps reduce overfitting and improve model robustness.

\subsection{Knowledge Distillation} \label{sec:distillation}

While we maintain an exemplar buffer to retain knowledge from previous tasks, its effectiveness is limited in the few-shot setting due to severe storage constraints. Following prior work~\citep{zhang2024continual}, we store only one exemplar per class. This storage constraint increases the risk of the model overfitting to the limited samples in the memory buffer. To alleviate this issue, following \cite{cao2020incremental, zhang2022efficient}, we employ a two-level distillation strategy comprising \emph{feature-level} and \emph{prediction-level} distillation.

\textbf{Feature-level Distillation.} To preserve the feature space learned in previous tasks, we distill knowledge from the model state after task $\mathcal{T}_{t-1}$ into the current model being trained on task $\mathcal{T}_t$. Let $f_{\theta_{t-1}}$ represent the model with its parameters frozen after completing task $\mathcal{T}_{t-1}$, and let $f_{\theta_t}$ be the current model. The distillation loss is defined as:
\begin{align}
\mathcal{L}_{fd} = 
\sum_{\xbm_i \in \widehat{\mathcal{D}}_t} \left(1 - \gamma (f_{\theta_{t-1}}(\xbm_i), f_{\theta_{t}}(\xbm_i) \right),
\end{align}
where $\gamma(\cdot,\cdot)$ is the cosine similarity function, used to measure the alignment between the representations from the previous and current models. This loss encourages the current model to generate feature representations consistent with the previous model, thereby mitigating catastrophic forgetting.

\textbf{Prediction-level Distillation.} The standard cross-entropy loss trains a model to maximize the probability of the correct label, which can bias the model towards the current task and accelerate the forgetting of earlier ones~\citep{wu2019large, cao2020incremental}. To mitigate this, we adopt prediction-level distillation, which transfers soft targets from the previous model to the current one. Let $\boldsymbol{\hat{p}}_{x,j}$ and $\boldsymbol{p}_{x,j}$ be the predicted probability distributions over the old classes from the previous and current models, respectively. The distillation loss is computed as the cross-entropy between these distributions:
\begin{align}
\mathcal{L}_{pd} &= 
\sum_{\xbm_i \in \widehat{\mathcal{D}}_t} \sum_{j \in \widehat{\mathcal{R}}_{t-1}} 
-\boldsymbol{\hat{p}}_{\xbm_i,j} 
\log \boldsymbol{p}_{\xbm_i,j}.
\end{align}
This distillation helps the model retain inter-class knowledge learned from previous tasks.

\textbf{Optimization Objective.} The final training objective for task $\mathcal{T}_t$ combines the standard classification loss with the router, label-contrastive, and distillation losses:
\begin{align}
    \min_{\theta_t, \phi}  
    \mathcal{L}_{ce} 
    + \alpha_\text{router} \mathcal{L}_{\text{router}} 
    + \alpha_\text{label} \mathcal{L}_{\text{label}} 
    +  \alpha_\text{fd} \mathcal{L}_{fd} 
    + \alpha_\text{pd} \mathcal{L}_{pd},
\label{eq:train_object}
\end{align}
where $\mathcal{L}_{ce}$ is the standard cross-entropy loss for the current task's data. The hyperparameters $\alpha_\text{router}$, $\alpha_\text{label}$, $\alpha_\text{fd}$, and $\alpha_\text{pd}$ balance the contribution of each component. During training, we update only the LoRA expert parameters $\theta_t$ and the event detector parameters $\phi$. A detailed overview of the training procedure is provided in Algorithm~\ref{alg:cap}.

%% file: content/algorithm.tex
\begin{algorithm}
\caption{Training Process for Task $\mathcal{T}_t$} \label{alg:cap}

\textbf{Input:} Training set $\widehat{\mathcal{D}}_t$, current event label set $\mathcal{R}_t$, previous model parameters $\theta_{t-1}$ and $\phi_{t-1}$.

\textbf{Output:}  Updated LoRA expert parameters $\theta_t$, event detector parameters $\phi_t$, accumulated memory buffer $\mathcal{\widehat{M}}_{t+1}$.

\begin{algorithmic}[1]
    \FOR{each event type $y \in \mathcal{R}_t$}
        \STATE Generate label descriptions for event type $y$.
        \STATE Encode descriptions using the base BERT model to obtain representations set $\mathcal{Z}_y$.
    \ENDFOR
    \FOR{epoch $e = 1$ to $E$}
    \FOR{each instance $\xbm_i \in \widehat{\mathcal{D}}_t$}
        \STATE Compute the CLS token representation $h^{\text{CLS}}_{\xbm_i}$ with a forward pass through the base BERT model.
        \FOR{layer $l = 1 \  \TO \ L$}
            \STATE Select expert indices $\boldsymbol{K}^l_{\xbm_i}$ using Equation~\eqref{eq:index_selection}.
            \STATE  Compute the layer output using the experts indexed by $\boldsymbol{K}^l_{\xbm_i}$ via Equation~\eqref{eq:update}.
        \ENDFOR
        \STATE Obtain final hidden representation $f_{\theta_t}(\xbm_i)$.
        \STATE Update model parameters $\theta_t$ and $\phi_t$ using the training objective in Equation~\eqref{eq:train_object}.
    \ENDFOR
    \ENDFOR
    \STATE Construct the new memory set $\mathcal{M}_t$.
    \STATE Update the memory buffer $\mathcal{\widehat{M}}_{t+1} = \mathcal{\widehat{M}}_{t} \cup \mathcal{M}_t$.
\end{algorithmic}
\end{algorithm}

%% file: tables/prompt.tex
\begin{table*}[ht]
\centering

\begin{tabular}{@{} p{\textwidth} @{}}
\toprule
 


\multicolumn{1}{l}{\textbf{Prompt to Generate Label Descriptions}} \\
\midrule
\textbf{Prompt: } I have a list of phrases called "Event". Write a complete, clear, and concise {\textbf{k}} description that explains the meaning, or general implication of the phrases.

Do not repeat words from the event list.

The output should follow the format:
Description: [...]

Each time you do it, try to change the phrasing, sentence structure, or word choice to create a different but meaningful version.

No additional titles, explanations, or quotes are needed in the output. 

Now, given the event: natural disaster, please generate 1 concise description.

\textbf{Output:}

Description: [An extreme environmental phenomenon causing considerable devastation and human suffering.]
\\
\bottomrule
\caption{Example generated label description.} 
\label{tab: instruct}
\end{tabular}

\end{table*}

%% file: content/experiment.tex
\section{Experiments} \label{sec: exp}



\subsection{Experimental Settings}

\textbf{Datasets.} We evaluate our proposed method on two benchmark datasets for FCED, maintaining consistency with prior work~\citep{cao2020incremental, zhang2024continual}:
\begin{itemize}
    \item \textbf{MAVEN} \citep{wang2020maven}: This dataset comprises 168 event types drawn from 118,732 instances. We use the preprocessed version provided by \cite{yu2021lifelong}. For the incremental learning setting, we partition the dataset into a sequence of five tasks, each containing four distinct event types. This setup corresponds to the \textit{4-way 5-shot} and \textit{4-way 10-shot} evaluation protocols, where each task provides 5 or 10 training instances per event type.

    \item \textbf{ACE-2005} \citep{walker2006ace}:  This dataset includes 33 event types. We adopt the preprocessed version provided by \cite{yu2021lifelong}. We partition the dataset into a sequence of five tasks, each containing two distinct event types. This results in \textit{2-way 5-shot} and \textit{2-way 10-shot} settings, with 5 or 10 training instances per event type for each task.
\end{itemize}

\input{tables/main_result}

\textbf{Baselines.} We evaluate our proposed method, LEAF, against a range of established continual learning approaches, including EWC \citep{kirkpatrick2017overcoming}, LwF \citep{li2017learning}, and ICaRL \citep{rebuffi2017icarl}, as well as prior CED techniques such as KCN \citep{cao2020incremental}, KT \citep{yu2021lifelong}, EMP \citep{liu2022incremental}, and HANet \citep{zhang2024continual}. Additionally, we adapt and compare LEAF with MoLE~\citep{wu2024mixture} for the FCED setting.

\textbf{Implementation Details.} Following prior work~\citep{zhang2024continual}, we employ the pre-trained BERT-base-uncased model~\citep{devlin2019bert} as our backbone. The training procedure comprises two stages: an initial fine-tuning phase on a base task, followed by an incremental learning phase under the few-shot conditions described above. For the LoRA experts, we set the rank to $r = 64$. All experiments were conducted on a single NVIDIA A100 GPU.

\textbf{Evaluation Metrics.} Consistent with prior CED research~\citep{cao2020incremental, zhang2024continual, dao2024lifelong}, model performance is evaluated using the F1-score. We report the average performance and standard deviation across five runs with different random seeds.

\subsection{Main Results}

The main experimental results on the MAVEN and ACE-2005 datasets are presented in Table~\ref{tab:main}. As shown, LEAF consistently outperforms conventional continual learning methods (EWC, LwF, and ICaRL) in the FCED setting. Compared to recent CED-specific methods such as KCN, KT, and EMP, LEAF also demonstrates substantial performance gains. Notably, on the final task, LEAF achieves an F1-score at least \textbf{10\%} higher on MAVEN and \textbf{24\%} higher on ACE-2005 relative to these baselines. This result highlights its effectiveness in mitigating catastrophic forgetting and its strong generalization in few-shot scenarios, which are two critical challenges in FCED.

Furthermore, LEAF exhibits significant improvements over the original MoLE architecture, which we adapted for this task. While MoLE's performance plateaus at an F1-score of approximately 40\%, LEAF consistently exceeds 50\% across all settings. Our method also outperforms HANet, a strong state-of-the-art baseline. On MAVEN, LEAF achieves an F1-score over \textbf{1\%} higher in both the 5-shot and 10-shot settings. On ACE-2005, the improvements are even more pronounced, with LEAF outperforming HANet by \textbf{6\%} and \textbf{3\%} in the 2-way 5-shot and 2-way 10-shot settings, respectively. These results underscore the effectiveness of LEAF’s MoE-based architecture in alleviating knowledge interference, a key limitation of the single-model approach used by HANet.

\subsection{Ablation Study}

\input{tables/augmented_des}

\textbf{Effectiveness of Each Component.} We conducted an ablation study on the ACE-2005 dataset to investigate the individual contributions of our framework's three key components: LoRA Experts, Knowledge Distillation, and Label Descriptions. The results are shown in Table~\ref{tab:label}. ``Baseline" is defined as the BERT model fine-tuned with a single LoRA module, without any of our proposed components. Removing all three components from our full model (LEAF) results in a significant performance drop, underscoring their collective importance. Introducing the LoRA Experts alone yields a notable \textbf{6.7\%} F1-score improvement on the final task in the 2-way 5-shot setting. This demonstrates their effectiveness in mitigating catastrophic forgetting by enabling specialized adaptation to new tasks and reducing knowledge interference. When Knowledge Distillation is added, the model achieves further substantial gains, improving performance by \textbf{8\%} in the 2-way 5-shot setting and the 2-way 10-shot setting over the baseline. This highlights the synergy between expert-based specialization and explicit knowledge transfer from previous tasks. Finally, integrating Label Descriptions provides an additional boost of up to \textbf{1.6\%} across both settings. This confirms their value in enhancing event type discrimination, particularly under data-scarce, few-shot conditions. Overall, these ablation results validate that each component provides a meaningful and complementary contribution to the final performance, and their joint integration leads to the state-of-the-art results achieved by LEAF.

\input{tables/num_of_label_des}

\textbf{Impact of the Number of Label Descriptions.} We investigated the impact of varying the number of label descriptions on model performance, with results presented in Table~\ref{tab:num_of_label_des}. While optimal performance is achieved with three descriptions per label, we find that using only a single description yields a result that is highly comparable. This finding has significant practical implications, given the computational and financial costs associated with generating descriptions using large language models. This result underscores LEAF's robustness and cost-effectiveness, making it a practical and efficient solution for low-resource scenarios.

\input{tables/num_expert}

\textbf{Impact of the Number of LoRA Experts.} We investigated the impact of varying the number of LoRA experts from 4 to 12 on model performance, with results presented in Table~\ref{tab:num_exp}. In the few-shot settings evaluated, optimal performance was achieved with four experts. This suggests that a small, well-utilized set of experts strikes an effective balance between model capacity and computational efficiency, providing sufficient expressiveness without introducing excessive parameters.
Conversely, increasing the number of experts beyond this optimal point leads to diminishing returns and, eventually, performance degradation. We attribute this decline to overfitting. With limited data, a larger number of experts introduces more parameters than can be effectively trained.  These findings highlight the importance of carefully tuning the number of experts to match the data constraints of the target application.

%% file: tables/main_result.tex
\begin{table*}[ht]
\centering
\resizebox{\textwidth}{!}{
\begin{tabular}{l|ccccc|ccccc}
\toprule
\multirow{2}{*}{Method} & \multicolumn{5}{c|}{MAVEN (4-way 5-shot)} & \multicolumn{5}{c}{MAVEN (4-way 10-shot)} \\
\cmidrule(lr){2-6} \cmidrule(lr){7-11}
& 1 & 2 & 3 & 4 & 5 & 1 & 2 & 3 & 4 & 5 \\
\midrule
EWC & 40.4$\pm$2.3 & 34.3$\pm$1.4 & 17.4$\pm$1.5 & 18.6$\pm$2.5 & 20.4$\pm$1.9 & 40.4$\pm$2.3 & 36.4$\pm$3.3 & 19.7$\pm$0.9 & 20.0$\pm$1.1 & 23.7$\pm$1.2 \\
LwF & 40.4$\pm$2.3 & 37.3$\pm$4.9 & 26.7$\pm$4.1 & 24.7$\pm$1.5 & 30.5$\pm$1.4 & 40.4$\pm$2.3 & 41.1$\pm$2.8 & 31.9$\pm$0.6 & 30.6$\pm$1.1 & 34.4$\pm$2.1 \\
ICaRL & 35.8$\pm$4.8 & 37.2$\pm$4.9 & 33.7$\pm$2.9 & 35.5$\pm$2.4 & 36.0$\pm$2.5 & 35.8$\pm$4.8 & 37.5$\pm$1.9 & 40.1$\pm$3.9 & 36.9$\pm$0.6 & 41.0$\pm$1.8 \\
KCN & 40.4$\pm$2.4 & 48.4$\pm$1.7 & 41.9$\pm$2.0 & 41.3$\pm$1.5 & 40.3$\pm$1.5 & 40.4$\pm$2.4 & 51.2$\pm$1.2 & 45.2$\pm$1.2 & 44.3$\pm$0.7 & 44.5$\pm$1.5\\
KT & 41.0$\pm$1.6 & 40.2$\pm$2.2 & 35.2$\pm$3.4 & 32.7$\pm$3.0 & 34.4$\pm$1.1 & 41.0$\pm$1.6 & 42.4$\pm$1.0 & 40.2$\pm$4.0 & 37.9$\pm$1.0 & 40.5$\pm$1.0 \\
EMP & 40.2$\pm$1.3 & 31.0$\pm$0.8 & 31.2$\pm$1.3 & 22.9$\pm$2.1 & 22.3$\pm$1.4 & 40.2$\pm$1.3 & 32.3$\pm$0.7 & 33.0$\pm$1.1 & 26.7$\pm$1.5 & 28.2$\pm$1.9 \\
MoLE &{42.1$\pm$1.1}&	49.9$\pm$1.0&41.0$\pm$1.2&40.9$\pm$1.7&40.4$\pm$1.0 & {42.1$\pm$1.1}&	48.4$\pm$1.1&41.7$\pm$1.5&39.5$\pm$2.4&	41.2$\pm$2.3 \\
HANet & 38.9$\pm$2.0 & \textbf{49.7$\pm$0.6} & \textbf{45.3$\pm$1.3} & {47.7$\pm$1.8} & 50.0$\pm$2.1 & {38.9$\pm$2.0} & \textbf{51.6$\pm$1.4} & {48.2$\pm$1.0} & 50.8$\pm$1.2 & 53.8$\pm$1.9 \\
\midrule
LEAF & \textbf{42.5$\pm$1.5} & 47.8$\pm$2.1 & {45.2$\pm$1.1} & \textbf{49.5$\pm$1.1} & \textbf{51.2$\pm$0.6} & \textbf{42.5$\pm$1.5} & {50.6$\pm$1.1} & \textbf{48.5$\pm$1.6} & \textbf{51.5$\pm$1.2} & \textbf{54.9$\pm$0.9} \\

\midrule
\multirow{2}{*}{Method} & \multicolumn{5}{c|}{ACE-2005 (2-way 5-shot)} & \multicolumn{5}{c}{ACE-2005 (2-way 10-shot)} \\
\cmidrule(lr){2-6} \cmidrule(lr){6-11}
& 1 & 2 & 3 & 4 & 5 & 1 & 2 & 3 & 4 & 5 \\
\midrule
EWC & \textbf{60.8$\pm$2.9} & 49.3$\pm$3.9 & 45.4$\pm$10.4 & 27.1$\pm$11.4 & 22.3$\pm$3.9 & \textbf{60.8$\pm$2.9} & 47.5$\pm$10.1 & 51.1$\pm$3.0 & 23.8$\pm$7.6 & 21.7$\pm$3.1 \\
LwF & \textbf{60.8$\pm$2.9} & 47.3$\pm$10.4 & 38.9$\pm$12.8 & 23.3$\pm$13.4 & 22.4$\pm$4.2 & \textbf{60.8$\pm$2.9} & 39.9$\pm$8.3 & 50.7$\pm$3.3 & 34.8$\pm$2.7 & 29.6$\pm$2.9  \\
ICaRL & 50.8$\pm$6.5 & 52.2$\pm$7.7 & 37.3$\pm$6.7 & 31.3$\pm$6.1 & 28.8$\pm$5.0 & 50.8$\pm$6.5 & 46.0$\pm$6.6 & 42.4$\pm$6.4 & 32.8$\pm$4.9 & 34.7$\pm$3.9 \\
KCN & \textbf{60.8$\pm$2.9} & 56.3$\pm$4.0 & 47.5$\pm$10.4 & 37.0$\pm$7.1 & 30.1$\pm$6.9 & \textbf{60.8$\pm$2.9} & 57.3$\pm$6.1 & 57.3$\pm$6.1 & 46.4$\pm$4.6 & 34.4$\pm$3.4  \\
KT & 53.1$\pm$2.2 & 42.5$\pm$2.3 & 33.9$\pm$2.9 & 38.4$\pm$6.8 & 31.2$\pm$3.9 & 53.1$\pm$2.2 & 59.1$\pm$1.7 & 50.0$\pm$5.1 & 46.0$\pm$5.3 & 28.5$\pm$2.9 \\
EMP & 54.7$\pm$1.4 & 40.4$\pm$1.9 & 24.3$\pm$3.8 & 27.1$\pm$8.4 & 22.5$\pm$6.0 & 54.7$\pm$1.4 & 37.2$\pm$7.3 & 19.6$\pm$6.7 & 32.9$\pm$4.7 & 24.1$\pm$6.6 \\
MoLE &	54.9$\pm$3.8 &54.4$\pm$3.2 &	42.7$\pm$3.2 &	39.0$\pm$5.0 &	39.9$\pm$4.8 & 54.9$\pm$3.8 &55.4$\pm$4.0	& 47.9$\pm$9.9 & 41.8$\pm$7.5 &	40.5$\pm$4.8\\
HANet & 60.2$\pm$2.7 & \textbf{62.8$\pm$2.4} & {45.0$\pm$3.1} & {45.0$\pm$2.1} & {49.4$\pm$1.3} & {60.5$\pm$2.5} & {61.3$\pm$4.1} & {58.4$\pm$2.1} & {59.3$\pm$1.3} & {58.2$\pm$0.9}\\
\midrule
LEAF & 57.2$\pm$2.2 & 61.7$\pm$1.5 & \textbf{49.3$\pm$2.0} & \textbf{50.8$\pm$2.3} & \textbf{55.5$\pm$2.5} & {57.2$\pm$2.2} & \textbf{63.3$\pm$2.3} & \textbf{62.2$\pm$1.5} & \textbf{60.0$\pm$0.8} & \textbf{61.6$\pm$2.5} \\
\bottomrule

\end{tabular}
}
\caption{\textbf{Overall Comparison.} Average F1-score (\%) on the MAVEN (4-way 5-shot and 4-way 10-shot) and ACE-2005 (2-way 5-shot and 2-way 10-shot) settings. Results are averaged over five runs. The best results are in \textbf{bold}.}
\label{tab:main}
\end{table*}

%% file: tables/augmented_des.tex
\begin{table*}
\centering
\resizebox{0.875\textwidth}{!}{
\begin{tabular}{l|ccccc|ccccc}
\toprule
\multirow{2}{*}{Method} & \multicolumn{5}{c|}{ACE-2005 (2-way 5-shot)} & \multicolumn{5}{c}{ACE-2005 (2-way 10-shot)} \\
\cmidrule(lr){2-6} \cmidrule(lr){7-11}
& 1 & 2 & 3 & 4 & 5 & 1 & 2 & 3 & 4 & 5 \\
\midrule
\ \ \ Baseline & 50.6 &58.7 &	46.8 &	46.5 &	43.0 &50.6 &	60.1&	56.5 &	48.8&	51.7 \\
+ LoRA Experts &	51.9 &	58.0&	38.8 &	48.3&49.7 &51.9 &	57.3 &	56.9 &	44.8 &	46.4 \\
+ Knowledge Distillation & {55.5} & {59.7} & {47.3} & {49.6} & {53.9} & {55.5} & {60.9} & {59.5} & {57.8} & {60.0} \\

+ Label Descriptions & \textbf{57.2} & \textbf{61.7} & \textbf{49.3} & \textbf{50.8} & \textbf{55.5} & \textbf{57.2} & \textbf{63.3} & \textbf{62.2} & \textbf{60.0} & \textbf{61.6} \\
\bottomrule
\end{tabular}
}
\caption{\textbf{Detailed analysis of the impact of the three proposed components.} We report the average F1-score (\%) on the ACE-2005 dataset. The best results are shown in \textbf{bold}.}

\label{tab:label}
\end{table*}

%% file: tables/num_of_label_des.tex
\begin{table*}
\centering
\resizebox{0.875\textwidth}{!}{
\begin{tabular}{l|ccccc|ccccc}

\toprule
\multirow{2}{*}{Number of Descriptions} & \multicolumn{5}{c|}{MAVEN (4-way 5-shot)} & \multicolumn{5}{c}{MAVEN (4-way 10-shot)} \\
\cmidrule(lr){2-6} \cmidrule(lr){7-11}
& 1 & 2 & 3 & 4 & 5 & 1 & 2 & 3 & 4 & 5 \\
\midrule
1 Description & {41.5} & \textbf{47.9} & \textbf{45.4} & {47.9} & {50.9} & {41.5} & {49.9} & {48.1} & \textbf{51.9} & {54.8} \\

3 Descriptions  & \textbf{42.1} & {47.8} & {45.2} & \textbf{49.5} & \textbf{51.2} & \textbf{42.1} & \textbf{50.6} & \textbf{48.5} & {51.5} & \textbf{54.9} \\

5 Descriptions & {39.9} & {47.6} & {45.3} & {50.0} & {51.2} & {39.9} & {49.5} & {47.9} & {51.6} & {54.5} \\
\midrule

\multirow{2}{*}{Number of Descriptions} & \multicolumn{5}{c|}{ACE-2005 (2-way 5-shot)} & \multicolumn{5}{c}{ACE-2005 (2-way 10-shot)} \\
\cmidrule(lr){2-6} \cmidrule(lr){7-11}
& 1 & 2 & 3 & 4 & 5 & 1 & 2 & 3 & 4 & 5 \\
\midrule
1 Description & 56.2 & 59.6 & {47.4} & {49.0} & {53.5} & {56.2} & {60.8} & {61.8} & {59.5} & {60.7} \\

3 Descriptions & \textbf{57.2} & \textbf{61.7} & \textbf{49.3} & \textbf{50.8} & \textbf{55.5} & \textbf{57.2} & 63.3 & \textbf{62.2} & \textbf{60.0} & \textbf{61.6} \\

5 Descriptions & 54.5 & 57.5 & 47.6 & 49.4 & 54.4 & 54.5 & \textbf{64.4} & 61.7 & 59.1 & 60.8 \\
\bottomrule

\end{tabular}
}
\caption{\textbf{Analysis of the number of label descriptions}. We report the average F1-score (\%) across varying numbers of descriptions on the MAVEN dataset and the ACE-2005 dataset. The best results are in \textbf{bold}.}

\label{tab:num_of_label_des}
\end{table*}

%% file: tables/num_expert.tex
\begin{table*}[ht]
\centering
\resizebox{0.875\textwidth}{!}{
\begin{tabular}{l|ccccc|ccccc}
\toprule
\multirow{2}{*}{Number of Experts} & \multicolumn{5}{c|}{MAVEN (4-way 5-shot)} & \multicolumn{5}{c}{MAVEN (4-way 10-shot)} \\
\cmidrule(lr){2-6} \cmidrule(lr){7-11}
& 1 & 2 & 3 & 4 & 5 & 1 & 2 & 3 & 4 & 5 \\
\midrule
{4 Experts} & \textbf{42.1} & \textbf{47.8} &\textbf{45.2} & \textbf{49.5} & \textbf{51.2} & \textbf{42.1} & \textbf{50.6} & \textbf{48.5} & \textbf{51.5} & \textbf{54.9} \\
{8 Experts} & {39.0} & {46.6} & {44.3} & {49.5} & {51.0} & {39.0} & {47.9} & {46.9} & {50.6} & {54.2} \\
{12 Experts} & {38.1} & {44.6} & {42.2} & {47.3} & {48.7} & {38.1} & {48.4} & {46.8} & {50.5} & {54.1} \\
\midrule


\multirow{2}{*}{Number of Experts} & \multicolumn{5}{c|}{ACE-2005 (2-way 5-shot)} & \multicolumn{5}{c}{ACE-2005 (2-way 10-shot)} \\
\cmidrule(lr){2-6} \cmidrule(lr){7-11}
& 1 & 2 & 3 & 4 & 5 & 1 & 2 & 3 & 4 & 5 \\
\midrule
{4 Experts} & \textbf{57.2} & \textbf{61.7} & \textbf{49.3} & \textbf{50.8} & \textbf{55.5} & \textbf{57.2} & {63.3} & \textbf{62.2} & \textbf{60.0} & \textbf{61.6} \\
{8 Experts} & {55.4} & {58.5} & {46.9} & {49.1} & {53.6} & {55.4} & \textbf{63.7} & {61.0} & {59.3} & {60.1} \\
{12 Experts} & {56.7} & {51.5} & {47.3} & {48.6} & {52.6} & {56.7} & {63.6} & {59.3} & {56.7} & {59.8} \\
\bottomrule
\end{tabular}
}
\caption{\textbf{Analysis of the number of LoRA experts}. We report the average F1-score (\%) across varying numbers of experts on the MAVEN dataset and the ACE-2005 dataset. The best results are in \textbf{bold}.}
\label{tab:num_exp}
\end{table*}

%% file: content/conclusion.tex
\section{Conclusion} \label{sec: conclude}

In this paper, we introduced \textbf{LEAF}, a novel expert-based framework designed to address key challenges in Few-shot Continual Event Detection. LEAF leverages multiple LoRA-based experts coupled with a semantic-aware routing mechanism to enable dynamic expert selection. This design promotes expert specialization across different input domains while mitigating knowledge interference between tasks. To enhance representation learning in low-resource settings, LEAF incorporates contrastive learning guided by semantically rich label descriptions. Additionally, a two-level knowledge distillation strategy is employed to prevent overfitting to the constrained memory buffer. Extensive experiments on standard FCED benchmarks demonstrate that LEAF consistently outperforms existing state-of-the-art methods, validating its effectiveness for continual event detection. 

Despite these strong results, our approach has certain limitations. LEAF's two-stage process for expert selection and final classification, used during both training and inference, increases computational overhead and prevents the model from being fully end-to-end. Future work could explore architectural improvements, such as a unified routing and classification mechanism, to enhance efficiency. Furthermore, LEAF currently relies on LoRA for parameter-efficient fine-tuning. Given recent advancements in this area, integrating diverse Parameter-Efficient Fine-Tuning strategies into our framework presents a promising direction for further improving performance and reducing computational costs.